\documentclass[aps,onecolumn,groupedaddress]{revtex4}

\usepackage[T1]{fontenc}
\usepackage{graphicx}
\usepackage{epstopdf}
\usepackage{amsmath}
\usepackage{bm}
\usepackage{float}
\usepackage[caption=false]{subfig}
\usepackage{adjustbox}
\usepackage[pdftex,hyperfigures,breaklinks,colorlinks,citecolor=blue,linkcolor=red]{hyperref} 
\usepackage{multirow}
\usepackage{gensymb}
\usepackage[normalem]{ulem} 
\usepackage{soul} 

\graphicspath{{Figures/}}

\renewcommand{\figurename}{{\bf Figure}}

\usepackage{longtable}
\usepackage{graphicx}
\usepackage{adjustbox}
\usepackage{multirow}
\usepackage[normalem]{ulem}
\usepackage{setspace}
\selectfont{}
\useunder{\uline}{\ul}{}
\begin{document}

\title{A novel data-driven algorithm to predict anomalous prescription based on patient's feature set}

\author{Qiongge Li$^{1}$, Jean Wright$^{1}$,  Russell Hales$^{1}$, Ranh Voong$^{1}$ and Todd McNutt$^{1}$\footnote{Corresponding author}, Email: tmcnutt1@jhmi.edu}

\affiliation{$^{1}$Department of Radiation oncology and molecular radiation sciences, Johns Hopkins University School of Medicine, Baltimore, MD 21205, USA}


\begin{abstract}
\begin{center}
\textbf{\abstractname}
\end{center}

Appropriate dosing of radiation is crucial to patient safety in radiotherapy. Current quality assurance depends heavily on a peer-review process, where the physician's peer review on each patient’s treatment plan including dose and fractionation. However, such a process is manual and laborious. Physicians may not identify errors due to time constraints and case load. We designed a novel prescription  anomaly detection algorithm that utilizes historical data from the past to predict anomalous cases. Such a tool can serve as an electronic peer who will assist the peer-review process providing extra safety to the patients.
In our primary model, we created two dissimilarity metrics, $R$ and $F$. $R$ defining how far a new patient's prescription is from historical prescriptions. $F$ represents how far away a patient's feature set is from that of the group with an identical or similar prescription. We flag prescription if either metric is greater than specific optimized cut-off values. We used thoracic cancer patients ($n=2356$) as an example and extracted seven features. Here, we report our testing f1 score, which is between 73\%-94\% for different treatment technique groups.  
We also independently validate our results by conducting a mock peer review with three thoracic specialists. Our model has a lower type II error rate compared to manual peer-review physicians.
Our model has many advantages over traditional machine learning algorithms, particularly that it does not suffer from class-imbalance. It can also explain why it flags each case and separate prescription and non-prescription-related features without learning from the data.  

\end{abstract}


\maketitle

\section{Introduction}\label{sec:introduction}

Radiotherapy (RT) is a complex process that requires careful quality assurance to ensure safe treatment delivery. One common safety concern is with errant or uncommon prescriptions (Rx) inadvertently being administered: excessively irradiating the patient can lead to injury or death. Meanwhile, under-irradiating may fail to mitigate cancer. Even though such events are rare, the impact of missing such errors could be catastrophic, and minor deviations result in sub-optimal treatment.

Peer review (PR) chart rounds are a significant component of the current quality assurance program in radiation oncology departments. In a study intended to evaluate the effectiveness of the PR process \cite{talcott2020blinded}, erroneous prescriptions and other anomalous cases were inserted into weekly rounds over nine weeks. Only 67\% of these anomalous prescriptions were detected by the physicians. Our goal is to present a data-driven algorithm to assist physicians by detecting anomalies automatically, which could potentially improve the patients' safety.

There is an increasing trend to study how machine learning (ML) tools can be used to augment medical professionals' decisions concerning diagnosis, treatment safety, and quality of patient care \cite{rajkomar2019machine, darcy2016machine}. Several pharmaceutical studies \cite{zhao2019cbowra,zhuo2020multiview,timonen2018electronic,hu2015detecting} have applied ML to find anomalous prescriptions but not tailored to RT. In RT, several studies \cite{CHANG2017S71,sipes2014anomaly,el2015detection,li2015treatment} have used ML to look at the treatment parameters to detect errors in treatment plans, but not focus on prescription error detection.

This work presents a multi-layer prescription anomaly detection tool that creates an automated, historical data-driven checkpoint to assist in PR. The tool's core utilizes a `distance model', which defines distance metrics between a new patient's features and prescriptions and those in a historical database. Prescription elements are the dose per fraction and the number of fractions prescribed to the target volume. Besides prescription features, there are other features such as diagnosis code, age at treatment, disease stage, treatment intent. Using a logical rule-based approach, the model will flag the new patient's prescription as anomalous if the distances fall outside certain optimized thresholds within a subgroup of similar patients.

\section{Methods}\label{sec:method}

\subsection{Data description}

We queried 15 years of cancer patients' radiotherapy treatment data (01/01/2006 – 07/13/2021) from MOSAIQ (a radiation oncology-specific electronic medical record). The data was comprised of 63768 individual treatment prescriptions which includes all the patients treated in the radiation oncology department of XXXX over the time span. 33 features related to patients' treatment information were extracted, including patient's age at treatment, diagnosis code, morphology code, treatment intent, techniques, energy, anatomic site, tumor stages and biomarkers. 

Prescription ($Rx$) data includes the number of fractions, dose per fraction, total dose, and accumulated total dose. Based on diagnostic codes, we grouped patients by disease site such as thoracic group, prostate group and etc. In this paper, we focus on thoracic subset (contains 11062 treatment plans). 
The Institutional Review Board of XXXX 
approved this study.

\subsection{Preprocessing and feature engineering}


Firstly, the raw data was split by technique. There were not enough samples to build models for the following treatment techniques: Intensity-modulated proton therapy (IMPT), Two-dimensional basic radiotherapy (2D), and Brachytherapy (Brachy), that we ruled out these techniques from our subsequent analysis. The techniques kept for later analysis are Three-dimensional conformal radiotherapy (3D), IMRT, and Stereotactic body radiotherapy (SBRT). 


 We removed highly rare energies used for each technique (i.e. x06fff is frequently used in SBRT but has never been used in 3D). Supplementary Table \ref{table:modality} is a summary of how frequently each energy was used for each technique.

Many further feature engineering steps were required to transform the columns of the data into a standardized form. Natural language processing (NLP) was used to converge many similar labels to a single values. In other cases irrelevant features needed to be removed. For example, Gleason scores were helpful for prostate cancer but irrelevant to the thoracic cancer.

In Supplementary Table \ref{table:diag_code}, we listed the diagnosis codes for our model and confirmed the completeness and appropriateness of this list for the model. Our current tool only included thoracic patients whose primary tumor site is the lung, heart, or esophagus. 
	
	
		We searched for re-plans and cone-down plans with their initials by finding the mismatch between the total and accumulated doses. 
		Because they are only 2.6\% of the total data points, we eliminated these patients' re-plan treatment along with their initial treatment. We also eliminated the cone-down plan records for the same reason.

    A number of additional checks were performed to future filter out atypical/strange data (i.e. samples with total dose do not match with fractions times the dose per fraction). Eventually, we acquired 2356 rows of records for thoracic. Supplementary Table \ref{table:dataexample} shows a sample post-processed feature-set.

%
\subsection{Model}\label{sec:data}

\begin{figure}[H]
	\centering
	\includegraphics[width=0.8\textwidth]{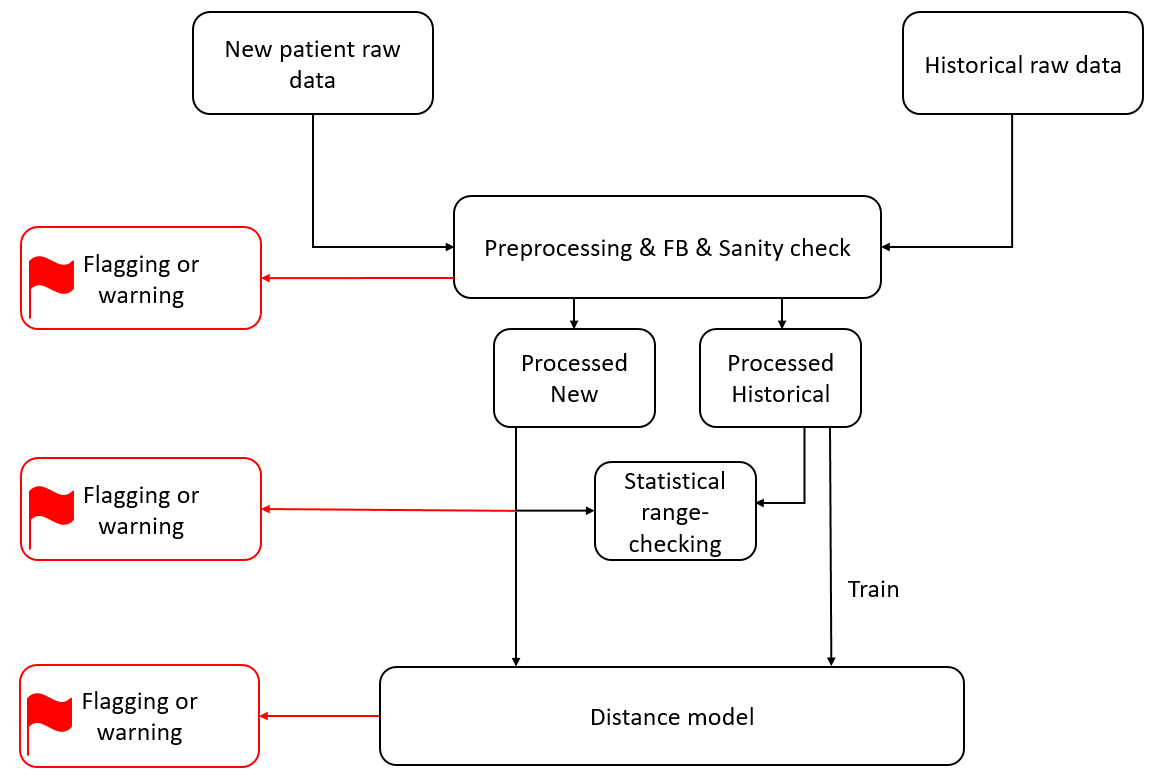}
	\caption{{\bf Model schematic.} The flow of data through the pipeline is illustrated as well as the different pipeline components and the different placess where flagging or warnings may occur.} 
	\label{fig:model_schema}
\end{figure}

 The essential idea of the model is to compare the new patient's prescriptions and other features to those in a historical database and to flag any any suspicious patterns because they have not been previously seen or are rare. In Fig. \ref{fig:model_schema} we can see the architecture of the model. The historical data and the new patient's are first processed as described in the previous section. Next, we explain the range checking and `distance model' components of the pipeline and under what circumstances the new patient's prescription will be flagged as a potential anomaly. 

\subsubsection{Statistical range-checking}\label{sec:method_level1}

 
 At this phase, we check whether the numerical values of the new patient's prescription are within normal historical ranges. We calculated the histogram of prescription features for a given technique over the historical database and fit a probability distribution curve to it. These distributions curves allow us to determine boundaries of normal prescriptions. Then, we flag any new patient's prescription if it falls outside these boundaries as shown in Supplementary Table \ref{table:boundary}. 
 


\subsubsection{Distance model}\label{sec:method_level3}

The distance model is designed to detect the following two types of prescription anomalies: the $Rx$ itself is atypical from the historical records (Type 1 anomaly) and there is a mismatch between $Rx$ and patients' other diagnostic features (Type 2 anomaly). 

The model defines a logical system that will flag the new patient if its `distance' from other patients in the historical database or specific groups of patients in the historical database is too large. In order to compare the new patient's prescription and other features with patients in the historical database, we need to define some pairwise and group level dissimilarity metrics. For this reason, we have defined two such distance metrics: a \emph{prescription distance} to indicate the distance in the prescription parameters, and a \emph{feature distance} to indicate distance within the remaining features included in the model.

The pairwise $Rx$ distance, $\rho_{Rx}(i,j)$ between the new patient, $i$, to any historical patient, $j$, in the database, is simply the Euclidean distance of the scaled prescription features,

\begin{equation}
	\rho_{Rx}(i,j) = \sqrt{(\widetilde{f}^{i}-\widetilde{f}^{j})^2 + (\widetilde{d}^{i}-\widetilde{d}^{j})^2}
	\label{eq:Rho_Rx}
\end{equation} 

where $\widetilde{f}$ and $\widetilde{d}$ is the min-max scaled fractions $f$ and dose per fraction $d$. 
\medskip

The pairwise feature distance, $g_{F}(i,j)$, between the new patient, $i$, and any historical patient, $j$, in the database, is the \textit{Gower} distance calculated overall features that are NOT prescription-related. The Gower distance \cite{gower1986metric} provides a simple way of computing dissimilarity when mixed numerical and categorical features are present. Numerical features contribute based on the absolute value of the difference divided by the range. In contrast, the dissimilarity is one for categorical features if they are different and zero if they are the same. Each feature in the Gower distance is given equal weight so that the Gower metric has a range on the interval [0,1].

\begin{equation}
	R(i,m) = \frac{1}{m}\sum_{j \in m- closest} \rho_{Rx}(i,j)
	\label{eq:R_function}
\end{equation}

\medskip
Similarly, we also define a ``closest-$n$ group distance'', $F(i)$, for all non-prescription-related features that apply the same formula but summing over $n$ pairwise Gower distances between the new patient, $i$ and patients, $k$, in the historical database. We restrict the sum to patients $k$ who have either \textit{the same prescription} as patient $i$ or who have minimal $Rx$ distance to patient $i$. For example, if $n=10$ and there are 12 patients with the same $Rx$ as patient $i$ in the historical, we select the lowest 10 Gower distances from this group of 12. If $n=20$, then first we would include all 12 terms $\rho_{Rx}(i,k)=0$ in the sum to compute $F$ and then sort over the next closest $Rx$ distance to find remaining terms similarly. We choose this metric because we expect features to be more similar when compared to others with the same (or similar) prescription.

\begin{equation}
	F(i,n) = \frac{1}{n}\sum_{k\in n-closest} g_{F}(i,k)    \qquad \text{where $n$ terms determined by sorting by $\rho_{Rx}(i,j)$ then by $g_{F}(i,k)$}
	\label{eq:F_function}
\end{equation}

In order to define thresholds that will define our cutoff for flagging, it is helpful to calculate some characteristic values of pairwise distances in the historical dataset. In this way, we can precisely define what we mean when we say two patients' features are similar or dissimilar. We can say they are dissimilar if their feature distance is much larger than the average historical pairwise distances for two patients with the same Rx. We compute the mean pairwise $Rx$ distance and the mean pairwise feature distance over all pairs of patients in the historical database to get a typical distance, $\theta$ and $\tau$, defined by

\begin{equation}
	\theta=\frac{1}{S(S-1)}\sum_{j,k}\rho_{Rx}(j,k)
	\label{eq:theta_bar}
\end{equation} 

\begin{equation}
	\tau=\frac{1}{S(S-1)}\sum_{j,k}g_{F}(j,k)
	\label{eq:tau_bar}
\end{equation}

where $S$ is the number of patients in the historical data base and, again,
$\rho_{Rx}(j,k)$, $g_{F}(j,k)$ are distances between a pair of historical patients $j$ and $k$.

Then, we pattern the thresholds as percentages of these characteristic values as follows:

\begin{equation}
	t_{Rx} = a\theta
	\label{eq:t_Rx}
\end{equation}

where $a$ is a model parameter to be determined by optimization. If $R > t_{Rx}$ then we flag it as an anomaly (Type 1).

Similarly, we define the feature threshold as a ratio of some characteristic values such as

\begin{equation}
	t_{F} = b\tau
	\label{eq:t_F}
\end{equation}

where $b$ is a model parameter. If $F > t_{F}$ then we flag as anomaly (Type 2).

In Supplementary Fig. \ref{fig:3dplot}, two different feature anomaly scenarios are depicted in a purely illustrative 3D feature space. In both cases, anomalies can be detected if far away from the $n$-group centroids belonging to their $Rx$. Note that in the diagram, the $n$-group centroids are determined by the data points on the surface of the $Rx$ cluster closest to each anomaly data point. In the {\bf a)}, the anomalies are isolated in the feature space, whereas in {\bf b)} a single anomaly is mismatched into an incorrect $Rx$ sector of the feature space.

\begin{figure}[H]
	\centering
	\includegraphics[width=0.55\textwidth]{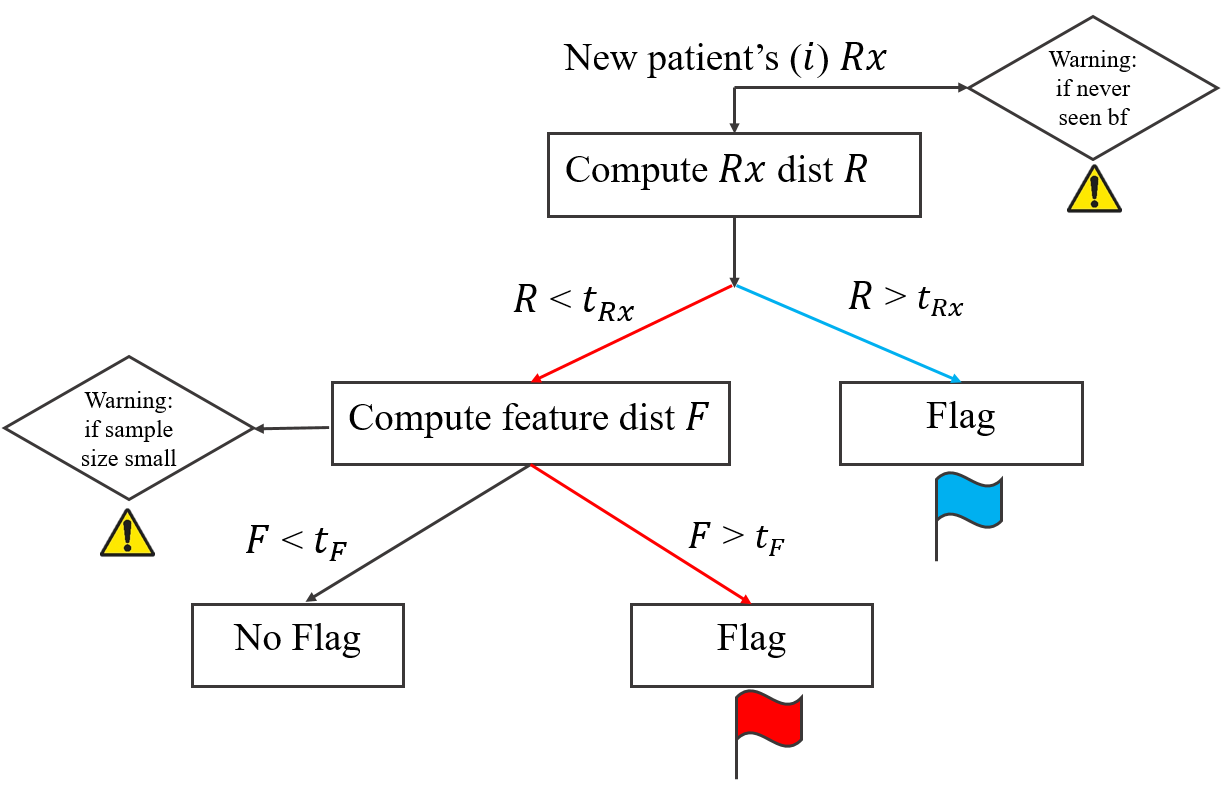}
	\caption{{\bf Model architecture}. We use dissimilarity metrics $R$ and $F$ to flag incoming new patient. If $Rx$ is uncommon ($R$ is greater than $t_{Rx}$), we flag it as blue. Otherwise, we compute feature distance $F$, if it is greater than cut off $t_{F}$, we flag it as red indicates that feature mismatches with its $Rx$. We can also give warnings as shown.}
	\label{fig:distance_model}
\end{figure}

The logic of the model is depicted in the decision tree shown in Fig. \ref{fig:distance_model}. The first step is to compute the closest-$m$ Rx group distance, $R(m)$, and flag if it is larger than some threshold $t_{Rx}$. If $R$ is too large, then the new patient's prescription is too dissimilar when considering other prescriptions in the historical database. If $R < t_{Rx}$ then we compute the closest-$n$ group feature distance considering only patients with the same $Rx$ as patient $i$ in the $F$ calculation. A warning is given if there are no $n$ patients in the historical database with the same prescription as the new patient, $i$.
If $F$ is more than some threshold $t_F$, we flag the new patient for the mismatch between the prescription and their other features, at least for the data in the historical database. The model was implemented in python.

\subsubsection{Model training}

We have four parameters in this model: $m$, $n$, $a$, $b$. In order to scale with the size of the historical dataset, the parameters $m$, $n$, are re-expressed as percentages of the historical training set size. Thus $m = \mu S$, where $S$ is the number of samples in the historical database per technique after subtracting a holdout set, and $r$ is the parameter we use for hyper optimization. Similarly, we define $n = \nu S$ and optimize of over the percentage $\nu$. Thus our final set of parameters for optimization are $\mu$, $\nu$, $a$ and $b$.

We used a parameter space search (grid-search) optimization to determine these parameters. The objective function for optimization was taken as the f1 score ($\frac{t_p}{t_p+\frac{1}{2}(t_p+f_n)}$; $t_p$ is true-positives and $f_n$ is false-negatives) over a training set that includes 10-30 simulated anomalies (SAs) and a similar number of non-anomalous patients.  Thus, the training set consists of SAs and holdout data from the historical database so that we have both positive (anomaly) and negative (not anomaly) classes in the test set.

Optimization through parameter space search was implemented with python \textit{hyperopt} module \cite{bergstra2013making}. Hyperopt uses the tree Parzen Estimator (TPE) to search the parameter space efficiently. Search intervals were defined based on the characteristic values $\theta$ and $\tau$ for parameters $a$ and $b$. Search intervals for the percentages $\mu$ and $\nu$ were constrained to be between 0 and 0.1, which confines the $m$, $n$-group dissimilarity metrics to 10\% of the historical database or lower for calculations of $F$ and $R$. The number of evaluations was set to 100 per each space search of the detection algorithm.

In order to reduce variance in the normal (not-anomaly) class, we averaged the results over random samplings of the non-anomalous holdout historical records. During this averaging, the anomaly class data points remained constant because we had a limited number of simulated anomalies available for training. This process was demonstrated in Supplementary Fig. \ref{fig:algorithm}.

\subsubsection{Synthesization of anomalies based on distribution}\label{sec:l4}

Creation of the anomalies is a time consuming task that requires careful examination of the historical database and identification of non-previously-occurring patterns between prescription and other features. We will illustrate the construction with some examples below. The main idea is to change the prescription of an existing record, or to change the other features of an existing record, in a way that creates a data point that is not typical of historical prescription-feature patterns. In this way we create \textit{a mismatch} between the prescription and the other features. This mismatch is verified by observing conditional distributions of features based on the given $Rx$ for each case. Thus we carefully check that the anomalies constructed are rare based on the historical conditional distributions. 

We must construct simulated anomalies that would be similar to those that could occur in the actual setting. We can obtain the correct parameters to generalize the model's application to the real world by carefully designing the anomalies. We expect to tune the model parameters to catch each of the simulated anomalies and flag them. 

Simulated anomalies were generated by switching the leading digit in the fractions with the leading digit in the dose per fraction or by varying several feature values randomly so that the resulting features do not match the prescription. In Supplementary Table \ref{table:anomaly_example}, we show four examples, marked A - D, where the original record is placed above its anomalous mutated form.  In example A, we switched the fractions (Fx) and dose per fraction (Dose/Fx) from 5 x 400 to 4 x 500. 5 x 400 is a common prescription in 3D thoracic treatment, having occurred 50 times in the historical database but not 4 x 500, which occurred only once.

The simulated anomalies were created in B and C by modifying other features and leaving the original prescription intact. For example, we changed the treatment intent from curative to palliative in case B and the age from 91 to 10. The prescription 5 x 1000 occurred 185 times in SBRT thoracic treatment but never occurred with palliative intent. Also, this Rx was never used in a pediatric patient (age under 21). Thus we varied the features in a way that created a mismatch between prescription and diagnostic features. In C, we mutated the diagnostic code from C34.30 to C15.9. Compared with the historical records, this Rx never treated the esophagus (which has a diagnostic code in the C15 series) and only was used to treat the lungs (C34 series). Also, we mutated the energy from x06 to x10, which never occurred for this Rx. 

In the last example, D, we simulated an anomaly by switching the technique label from 3D to IMRT so that effectively all the features are mismatched. 10 x 300 is a common Rx in both 3D and IMRT. The feature sets are pretty distinct because in 3D, the energy that comes with this Rx is usually 15x, but 15x rarely occurs in historical IMRT cases. 

It should be noted that this approach to simulating anomalies is purely data-driven and based on deviations from past historical patterns. The anomaly creation process was done by authors with no clinical information (MDs were excluded from this process).  


\section{Results}\label{sec:results}

Here, we provide illustrative results from running the distance model. In Supplementary Fig. \ref{fig:pairwise_dist}, we plotted the histograms of $Rx$ and feature distances from historical DB. We can see that the Rx distances of zero or 0.2 are particularly common which reflects the fact that many patients in the dataset have the same or similar prescriptions. The feature distances are more varied, and display characteristic spikes associated with the categorical differences (see Supplementary Fig. \ref{fig:pairwise_dist} caption for further explanation).

As discussed in Sec. \ref{sec:l4}, there are several ways in which we synthesized anomalies. We present the in-sample training results for the $Rx$ switched (see Example A in Supplementary Table \ref{table:anomaly_example}) type of SAs in Table \ref{table:parameter}. The $S$ column refers to the number of records in the historical database, $a$,$b$ are the parameters multiplying $\theta$ and $\tau$ respectively, and $\mu=\frac{m}{S}$ and $\nu=\frac{n}{S}$ are the parameters $m$ and $n$ expressed as percentages of $S$. $s_a$ refers to the number of anomalies in the training set, whereas $s_n$ refers to the number of normal \textit{holdout} historical samples in the training set. Note that the holdout set $s_n$ is not used to compute $\theta$ or $\tau$.

 \begin{table}[H]
	\caption{Parameters and model performance scores. }
	\label{table:parameter}
	\centering
	\begin{tabular}{|l|c|c|c|c|c|c|c|c|c|c|c|}
		\hline
		& Technique                      & a     & b     & $\nu$ & $\mu$ & $\tau$ & $\theta$ & $f1$ & $s_n$ & $s_a$ & $S$  \\ \hline
		\multirow{3}{*}{\begin{tabular}[c]{@{}l@{}}Rx \\ switched \\ SAs\end{tabular}} &
		3D &
		0.449 &
		1.632 &
		0.012 &
		0.018 &
		0.581 &
		0.206 &
		0.98 $\pm$ 0.03 &
		20 &
		10 &
		509 \\ \cline{2-12} 
		& IMRT & 0.265 & 0.979 & 0.025 & 0.014 & 0.543  & 0.261    & 0.89 $\pm$ 0.01                & 20    & 10    & 1153 \\ \cline{2-12} 
		& SBRT & 1.631 & 1.838 & 0.047 & 0.014 & 0.501  & 0.142    & 0.98 $\pm$ 0.03                & 20    & 10    & 704  \\ \hline
		\multirow{3}{*}{\begin{tabular}[c]{@{}l@{}}Feature\\ switched\\ SAs\end{tabular}} &
		3D &
		0.056 &
		0.797 &
		0.021 &
		0.019 &
		0.581 &
		0.206 &
		0.84 $\pm$ 0.02 &
		20 &
		20 &
		509 \\ \cline{2-12} 
		& IMRT & 0.286 & 0.802 & 0.023 & 0.038 & 0.543  & 0.261    & 0.84  $\pm$ 0.01               & 20    & 20    & 1153 \\ \cline{2-12} 
		& SBRT & 0.307 & 0.584 & 0.017 & 0.029 & 0.501  & 0.142    & 0.90 $\pm$ 0.03                & 20    & 20    & 704  \\ \hline
		\multirow{3}{*}{\begin{tabular}[c]{@{}l@{}}In-sample\\ (both types \\ of SAs)\end{tabular}} &
		3D &
		0.010 &
		0.717 &
		0.010 &
		0.037 &
		0.581 &
		0.206 &
		0.84 $\pm$ 0.01 &
		30 &
		30 &
		499 \\ \cline{2-12} 
		& IMRT & 1.401 & 0.805 & 0.025 & 0.014 & 0.543  & 0.261    & 0.86 $\pm$ 0.01                & 30    & 30    & 1143 \\ \cline{2-12} 
		& SBRT & 1.926 & 0.465 & 0.01  & 0.075 & 0.501  & 0.142    & 0.91 $\pm$ 0.03                & 30    & 30    & 694  \\ \hline
		\multirow{3}{*}{\begin{tabular}[c]{@{}l@{}}Out-of-sample\\ (both types\\ of SAs)\end{tabular}} &
		3D &
		0.010 &
		0.717 &
		0.010 &
		0.037 &
		0.580 &
		0.200 &
		0.941 &
		10 &
		8 &
		529 \\ \cline{2-12} 
		& IMRT & 1.401 & 0.805 & 0.025 & 0.014 & 0.544  & 0.273    & 0.727                          & 10    & 10    & 1173 \\ \cline{2-12} 
		& SBRT & 1.926 & 0.465 & 0.010 & 0.075 & 0.503  & 0.141    & 0.875                          & 10    & 7     & 724  \\ \hline
	\end{tabular}
\end{table}

The $f1$ score was computed by averaging over 50 trials of random samples of the not-anomaly holdout set $s_n$. We found $f1$ scores of 0.98 for 3D, 0.89 for IMRT, and 0.98 for SBRT, where the error bars run between 2-5\%. For the feature switching generated SAs, we found $f1$ scores of 0.84 for 3D, 0.84 for IMRT, and 0.90 for SBRT with similar error bars, as shown in Table \ref{table:parameter}.

Next, we ran the model on a training set combining both $Rx$ switched and feature switched SAs. We found that the resulting $f1$ scores for the combined training set lie in between the scores for the training sets where each type of anomaly was considered separately. This makes sense intuitively. We report the results and parameters in Table \ref{table:parameter}. Because the standard deviation is small, we choose any run as our final parameters. Note that $\tau$ or $\theta$ varies slightly because of the different historical holdout samples.

 Out-of-sample results are obtained by running the distance model with the same parameters that were found during optimization over the training set, on the new unseen test set. E.g., in the test set, both the normal `non-anomalous' test records and the anomalous test records are previously unknown to the distance model. 
 

We used a separate, recent data set (01/01/2021 - 07/14/2021) to select samples for our out-of-sample testing non-anomalous class data. We used all of the samples during this time period for the 3D and SBRT, each containing ten samples. We selected 10 of the most typical cases out of the 24 IMRT samples from this time period as our testing normal class. For the out-sample case, the historical data set (from 01/01/2006 - 12/31/2021) is still an important input into the model, however, no samples are drawn from it for prediction. We then created a new set of SAs for each technique using several construction methods and verified the anomalous class status by looking at the conditional feature distribution after switching/changing features.

 We report the out-of-sample distance model results in Table \ref{table:parameter}. We can see that comparing the out-of-sample performance to the in-sample, the out-of-sample is worse for IMRT and SBRT but better for 3D.

A beneficial feature of the distance model is that not only do we get the model prediction for each of the test records, but we also get an explanation of why each prediction was made. By looking at the values of $R$, $F$, $t_F$ and $t_{Rx}$ we can immediately see the reason why a sample was flagged or not flagged, as shown for example in Table \ref{table:example_predictions}, where each row represent a testing patient. 

\begin{table}[H]
	\caption{Prediction examples}
	\label{table:example_predictions}
	\centering
	\begin{tabular}{|c|c|c|c|c|c|c|c|c|c|c|c|c|c|c|c|}
		\hline
		Fractions & Dose/Frac & Tech & Modality & Intent & ICD10 & ICDO & Age & Truth & Pred & Type & R & $t_{Rx}$ & F & $t_f$ & Counts \\ \hline
		4 & 1200 & SBRT & x10fff & palliative & C15.6 & 87203 & 49 & 1 & 1 & 2 & 0.00 & 0.27 & 0.56 & 0.23 & 417 \\ \hline
		4 & 1200 & SBRT & x06fff & curative & C34.12 & & 61 & 0 & 0 & & 0.00 & 0.27 & 0.21 & 0.23 & 417 \\ \hline
		4 & 500 & 3D & mixed photon & & C34.90 & 80463 & 76 & 1 & 1 & 1 & 0.13 & 0.00 & & & 1 \\ \hline
	\end{tabular}
\end{table}

The `Truth' column refers to whether the data point is actually an anomaly or not (1 indicates anomaly; 0 indicates normal). The `Pred' column is the prediction by the model, where again 1 indicates anomaly and 0 indicates not. The first row was predicted by the model to be a `Type 2' anomaly, which means the feature distance is large for this patient compared to the historical database (the new patient's feature sets do not match well with the population who received the same prescription in the past). $R$ is 0 which is below the cut off $t_{Rx}$ but $F$ is larger than the cut off value $t_{F}$. $R$ is zero is because this Rx has been seen in the historical database (The `Counts' column indicates it has been seen 417 times previously). Observation of historical distributions shows that the energy 10fff was never previously used for the $Rx$ prescription $4 \times 1200$, and this prescription was never used to treat an esophagus diagnosis either. This, again, shows that being a ``common'' prescription cannot promise being ``normal'' or not an error.

 The second row is a normal patient in the database where the feature sets match well with the historical record. Therefore $R$ and $F$ are both smaller than cut-off values. In the third row, we show a switching anomaly, the original prescription was 5 x 400, but we switched it to 4 x 500. This leads to a large $Rx$ distance $R$, making the model predict it as Rx anomaly (`Type 1'). This is consistent because 4 x 500 almost never appears in the historical database (Counts = 1). This again, shows that our model has the ability to not only predict anomalies but also to \textit{explain} each prediction.

\subsection{Mock peer review (PR)}

\subsubsection{Model's performance vs. individual's performance}
\begin{figure}[H]
	\centering
	\includegraphics[width=0.7\textwidth]{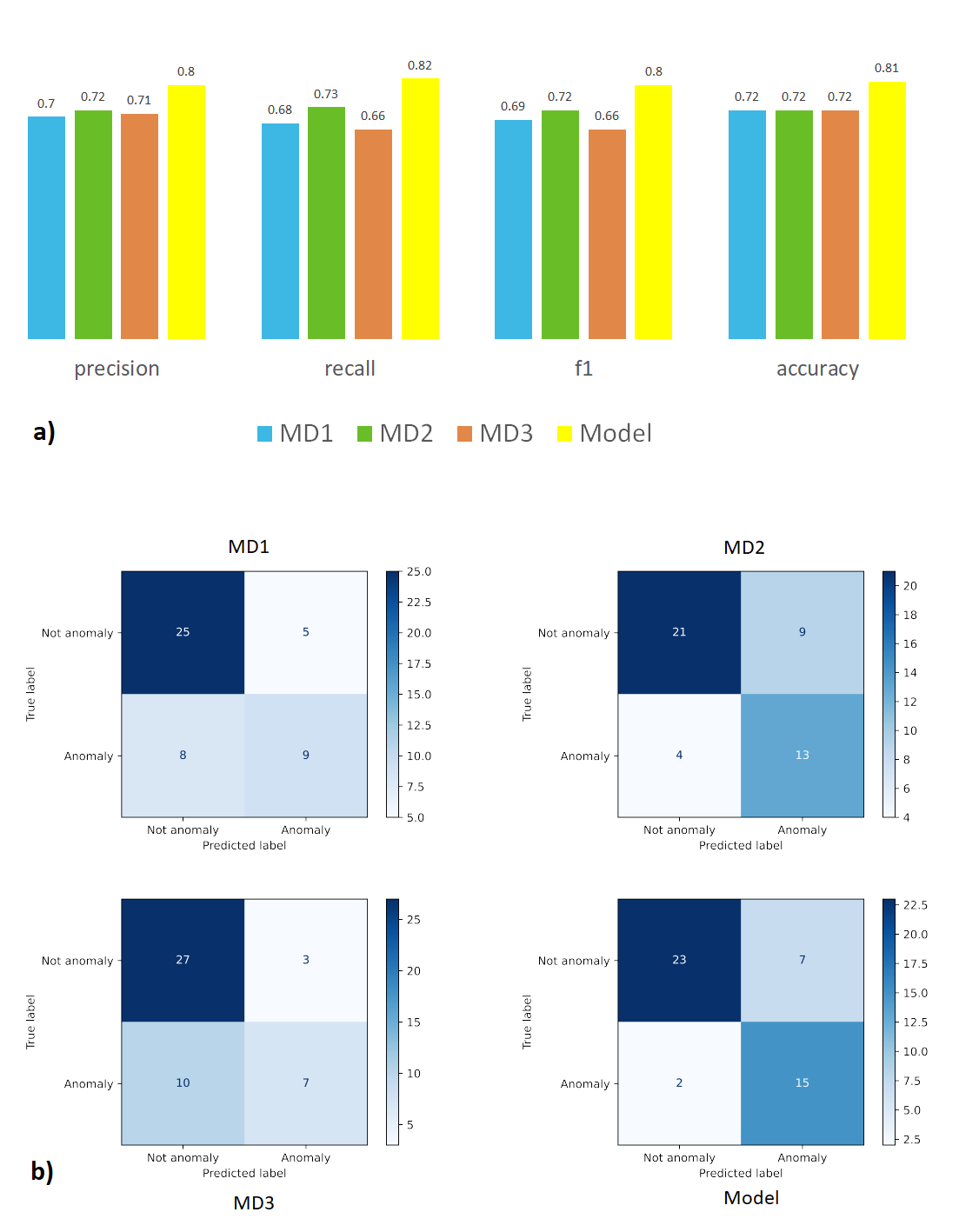}
	\caption{{\bf Performance}. {\bf a)} Performance (macro average of metrics) of physicians vs. model. The blue, green and orange bar indicates each physician's performance and the yellow bar is the model performance. We can see that the model's precision, recall, f1 and accuracy scores are all compatible with the physicians suggesting that the model can serve a role as a digital peer.
		{\bf b)} Confusion matrix. The model has the lowest false-negative rate, suggesting that the model is more conservative than all the physicians, in deciding whether a case should be considered an anomaly.} 
	\label{fig:confusion_matrix}
\end{figure}

In order to independently validate our results, we conducted a mock RP. Three radiation oncologists with more than ten years of experience treating thoracic patients were each asked to independently label a sample dataset containing 17 anomalies and 30 normals (a subset randomly selected from out-of-sample testing data). The results of the physicians, side-by-side with the model results, are shown in Fig. \ref{fig:confusion_matrix} {\bf a)}. The performance was evaluated by calculating precision, recall, f1 and accuracy. 
Additionally, confusion matrices for the physicians (MDs) and the model are shown in Fig. \ref{fig:confusion_matrix} {\bf b)} which gives a breakdown of the different type I and type II errors made by each physician and the model.

In Supplementary Table \ref{table:cases}, we show some specific examples of cases.

\subsubsection{Time analysis}
To get a sense of the time and effort spent by each physician on the mock PR, we asked each physician to note the time spent on the review. MD2 spent 18 minutes identifying the errors and 12 minutes writing out the rationale. MD1 spent a total of 11 minutes both identifying the errors and writing out the rationale for their decisions. The model running time for a single testing sample is about 1s and the model training time is several days. However, one only needs to train the model once. The training time is proportional to the number of evaluation points in the grid space, the number of runs to average the $f1$ score and the number of data samples.

\subsubsection{Model's performance vs. physician group's performance}
In the RP, physicians can discuss each case and combine their knowledge to form a consensus about the correctness of a prescription for each case under review. Thus, besides comparing our model's performance against each physician individually, we also compare it with the group consensus. We consider a best and worst-case scenario from joining MDs. In the best case, the consensus is correct if any MD was correct; in the worst case, the RP selects an incorrect decision if any MD was wrong. We would expect actual performance of RP in the real clinical setting would lie in between. 

\begin{figure}[H]
	\centering
	\includegraphics[width=0.7\textwidth]{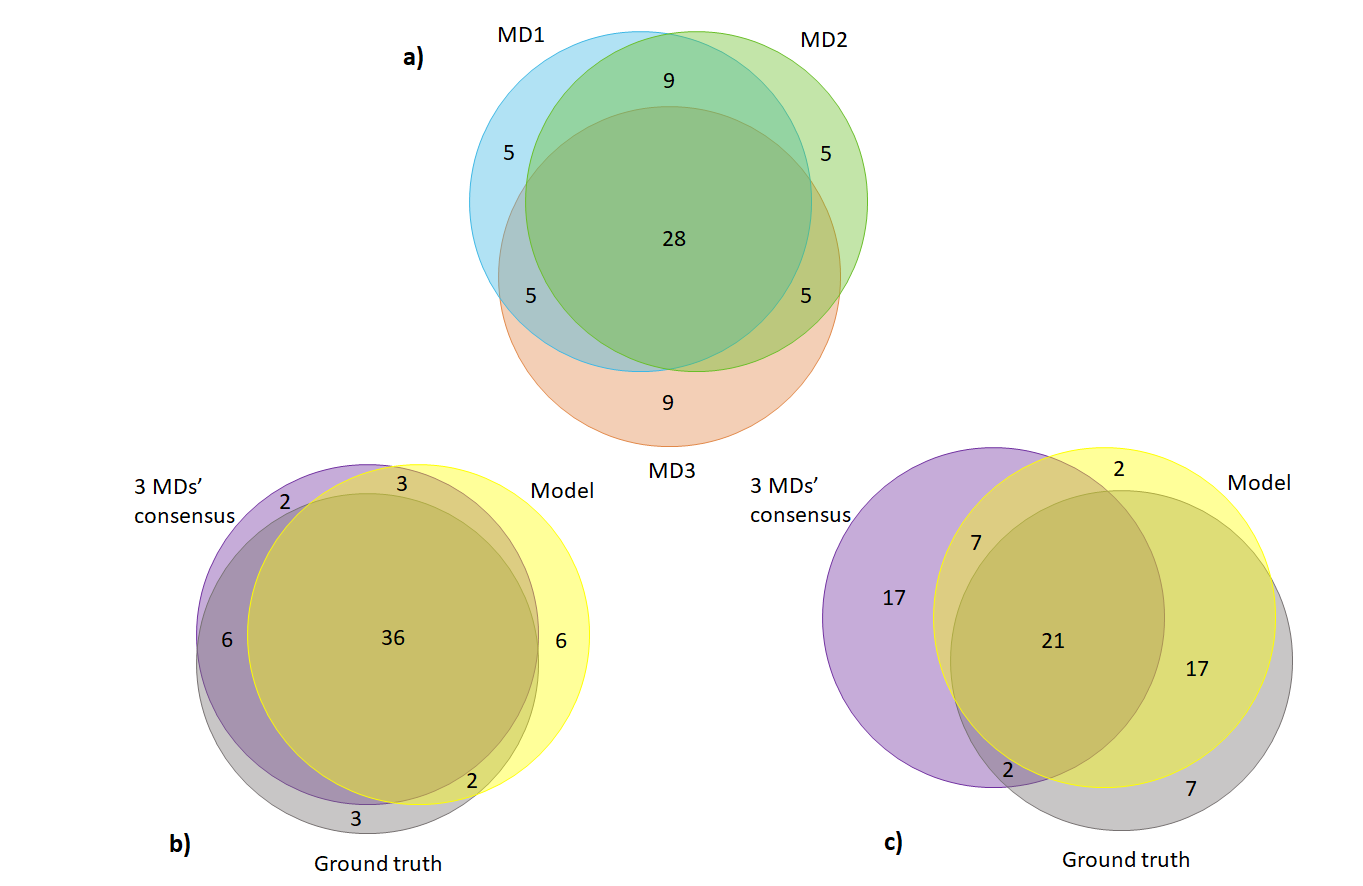}
	\caption{{\bf Overlap of agreement between MDs, the model and the ground truth}. Panel {\bf a)} shows the overlap of agreement between the three MDs on decisions of whether to flag or not to flag a particular case. Panel {\bf b)} shows the best-case scenario from the PR and {\bf c)} shows the worst-case scenario. } 
	\label{fig:overlaps}
\end{figure}

The results of such a worst and best case scenario are displayed in Fig. \ref{fig:overlaps} as well as the overlap diagrams of agreement for each individual MDs. Note that the numbers in the Venn diagrams do not distinguish between anomalous or non-anomalous class. Any overlap regions with the ground truth set correspond to correct decisions, any decisions outside the ground truth set correspond to incorrect decisions. We can see that, in the worst-case, panel {\bf c)}, the model outperformed the consensus by missing 9 (2 + 7) cases rather than 24 (17 + 7) cases by the consensus. However, the real question is whether the model is still better than the best-case's consensus, {\bf b)}? The answer is no. The model missed 9 (3 + 6) cases, while the best-case consensus missed 5 (2 + 3) cases. Our model's performance is in between the best and worst-scenario, but closer to the former. The overlapping regions/agreements indicates that the model independently agreed with physician's knowledge.  

We should not interpret these results to suggest that the model under-performed or out-performed the MDs in the mock PR. Instead, we suggest that the model be considered an additional ``digital peer reviewer'' to complement the MDs. Under these circumstances, the distance model has promise as a validation tool to check for prescription errors since the model caught anomalies that the physicians overlooked.

\section{Discussion}\label{sec:discussion}

One of the advantages of this model compared to a supervised learning (SL) model is that it does not present any problem with class imbalance. This is because the distance model is not an SL model in the traditional sense and instead relies on distances between historical data and the test set to define outcomes.
When comparing the model's performance versus the physician's performance we note that even with the same level of performance, the model is still valuable because it is a fully automated process that does not require valuable physician time and provides an additional safety check.


This approach has advantages over SL models that are not a good fit for anomaly detection. It is possible that similar results could be constructed with existing anomaly detection techniques such as isolation forest or $k$-means. However, it is unclear how to separate the prescription features from the diagnostic/other features, as discussed in the Introduction. 

An alternative method in this context \cite{sharabi2012automated} calculates the conditional probability of the $Rx$ conditioned on the features and threshold for the rarity. We could have expanded this idea by calculating every conditional probability of the features on the $Rx$, or features on other features and threshold for rarity for the same Rx. However, a major drawback of this approach is that it involves many condition-by-condition checking. In contrast, our approach is simpler where we saved many efforts in avoiding checking case-by-case. 





\subsection{Limitations}\label{sec:limit}

We are limited by the number of informative features that we can build and the available data. Lack of features limits our ability to make predictions, and lack of data increases the variance in whatever predictions we can make. To increase the number of features, NLP would be needed to encode the physician's notes into a vector, which we can calculate pairwise distances over. More data could be obtained by merging datasets with other institutions. 

Another major limitation is the difficulty of constructing or obtaining anomaly data. It is challenging to make realistic anomalies because they are rare and unexpected by their nature, so creating a set of anomalies that fully samples the space of possibilities is a significant challenge. It would be helpful to have more anomalous data for validation. The relatively small number of anomalies limits the scope of our findings. 


\section{Conclusions}\label{sec:conclusion}

 In this report, we have provided a proof-of-concept for an anomaly detection pipeline for prescription in radiotherapy. Our results show that the distance model and connected pipeline can predict with good accuracy for anomalies that are constructed according to the methods described in Sec. \ref{sec:l4}. The model showed promise and was evaluated favorably in the mock clinical setting where its predictions agreed independently with physicians' knowledge and, in some cases, out-performed the physicians. Our approach has focused on a custom decision tree rule-based anomaly detection logic that creates its own definitions of ``dissimilarity'' between historical patient data. These dissimilarities are incorporated into a pipeline with novel decision tree logic that is a potentially useful and novel approach to prescription anomaly detection in the RT setting.


\section{Author contributions statement}
Q.L.: Methodology and Software; Data analysis; Visualization; Writing - Original Draft. 
R.V., R.H.: Performed Mock peer review; Discussion. 
J.R.: Performed Mock peer review; Discussion; Funding acquisition. 
T.M.: Supervision; Funding acquisition. 
All authors reviewed and edited the manuscript.

\section{Competing interests}
TM: Oncospace Inc. 
All other authors declare no competing interests.
\section{Data and code availability}
This data set and code can be available by contacting the corresponding author. 

\clearpage
\section*{REFERENCES}
\bibliographystyle{unsrtnat}

\bibliography{peer_review}

\setcounter{section}{0}
\setcounter{figure}{0}

\renewcommand{\figurename}{{\bf Supplementary Figure}}
\renewcommand{\tablename}{{\bf Supplementary Table}}

\setcounter{table}{0}
\setcounter{figure}{0}
\renewcommand{\thetable}{S\arabic{table}}
\renewcommand\thefigure{S\arabic{figure}}
\renewcommand{\theHtable}{Supplement.\thetable}
\renewcommand{\theHfigure}{Supplement.\thefigure}

\clearpage
\onecolumngrid

\section*{\bf {\Large Supplementary Information}}
\subsection{Supplementary tables}
\begin{table}[H]
	\caption[Energy used]{Energy used in different techniques}
	\label{table:modality}
	\centering
	
	\begin{tabular}{|c|c|c|c|}
		\hline
		\textbf{Energy} & \textbf{3D} & \textbf{IMRT} & \textbf{SBRT} \\ \hline
		x06               & 293          & 1896           & 277           \\ \hline
		x06FFF            & 0           & 71            & 615           \\ \hline
		x10               & 77          & 56            & 1             \\ \hline
		x10FFF            & 0           & 1             & 0             \\ \hline
		x15               & 521         & 94             & 8             \\ \hline
		Mix Photon        & 144          & 76            & 1             \\ \hline
		Mix Mode          & 1           & 0             & 0             \\ \hline
	\end{tabular}
	
\end{table}

\begin{table}[H]
	\caption[diagnosis code included in the model]{Diagnosis codes included in the thoracic model.}
	\label{table:diag_code}
	\centering

	\begin{tabular}{|l|l|}
		\hline
		\textbf{Diagnosis code (ICD10)} & \textbf{Description of the code}                              \\ \hline
		C15.3                           & Malignant neoplasm of upper third of esophagus                \\ \hline
		C15.4                           & Malignant neoplasm of middle third of esophagus               \\ \hline
		C15.5                           & Malignant neoplasm of lower third of esophagus                \\ \hline
		C15.9                           & Malignant neoplasm of esophagus, unspecified                  \\ \hline
		C33                             & Malignant neoplasm of trachea                                 \\ \hline
		C34.00                          & Malignant neoplasm of unspecified main bronchus               \\ \hline
		C34.01                          & Malignant neoplasm of right main bronchus                     \\ \hline
		C34.02                          & Malignant neoplasm of left main bronchus                      \\ \hline
		C34.10                          & Malignant neoplasm of upper lobe, unsp bronchus or lung       \\ \hline
		C34.12                          & Malignant neoplasm of upper lobe, left bronchus or lung       \\ \hline
		C34.2                           & Malignant neoplasm of middle lobe, bronchus or lung           \\ \hline
		C34.30                          & Malignant neoplasm of lower lobe, unsp bronchus or lung       \\ \hline
		C34.31                          & Malignant neoplasm of lower lobe, right bronchus or lung      \\ \hline
		C34.32                          & Malignant neoplasm of lower lobe, left bronchus or lung       \\ \hline
		C34.80                          & Malignant neoplasm of ovrlp sites of unsp bronchus and lung   \\ \hline
		C34.81                          & Malignant neoplasm of ovrlp sites of right bronchus and lung  \\ \hline
		C34.82                          & Malignant neoplasm of ovrlp sites of  left bronchus and lung  \\ \hline
		C34.90                          & Malignant neoplasm of unsp part of unsp bronchus or lung      \\ \hline
		C34.91                          & Malignant neoplasm of unsp part of  right bronchus or lung    \\ \hline
		C34.92                          & Malignant neoplasm of unsp part of left bronchus or lung      \\ \hline
		C37                             & Malignant neoplasm of thymus                                  \\ \hline
		C38.1                           & Malignant neoplasm of anterior mediastinum                    \\ \hline
		C38.2                           & Malignant neoplasm of posterior mediastinum                   \\ \hline
		C38.3                           & Malignant neoplasm of mediastinum, part unspecified          \\ \hline
		C38.4                           & Malignant neoplasm of pleura                                  \\ \hline
		C38.8                           & Malignant neoplasm of ovrlp sites of heart, mediastinum and pleura \\ \hline
		C45.0                           & Mesothelioma of pleura                                        \\ \hline
		C77.1                           & Secondary and unsp malignant neoplasm of intrathorac nodes    \\ \hline
		C78.00                          & Secondary malignant neoplasm of unspecified lung              \\ \hline
		C78.01                          & Secondary malignant neoplasm of right lung                    \\ \hline
		C78.02                          & Secondary malignant neoplasm of left lung                     \\ \hline
		C78.1                           & Secondary malignant neoplasm of mediastinum                   \\ \hline
		C78.2                           & Secondary malignant neoplasm of pleura                        \\ \hline
		D15.0                           & Benign neoplasm of thymus                                     \\ \hline
		E85.8                           & Other amyloidosis                                             \\ \hline
		R91.1                           & Solitary pulmonary nodule                                     \\ \hline
	\end{tabular}

\end{table}
\clearpage
	\begin{table}[H]
	\caption[]{Data row example}
	\label{table:dataexample}
	\centering
	\begin{tabular}{|c|c|c|c|c|c|c|c|}
		\hline
		\multicolumn{2}{|c|}{Rx-related features} & \multicolumn{6}{c|}{None-Rx-related features}                        \\ \hline
		Fx                & Dose/fx               & Age at Tx & Technique & Energy & Intent     & ICD10 code & Morphology code \\ \hline
		4                 & 1200                  & 63        & SBRT      & x06    & palliative & C34.10     & 81406     \\ \hline
	\end{tabular}
\end{table}

\begin{table}[H]
	\caption[Energy used]{Boundaries of BED, Fractions and Dose per fraction for different techniques}
	\label{table:boundary}
	\centering
	\begin{tabular}{|c|c|c|c|c|c|c|}
		\hline
		& \textbf{min(BED)} & \textbf{max(BED)} & \textbf{min(Fractions)} & \textbf{max(Fractions)} & \textbf{min(Dose/Fractions)} & \textbf{max(Dose/Fractions)} \\ \hline
		\textbf{3D}   & 16400  & 292800 & 1 & 35 & 150  & 850  \\ \hline
		\textbf{IMRT} & 24000 & 497000 & 7 & 47 & 150 & 700  \\ \hline
		\textbf{SBRT} & 82000 & 903000 & 1 & 5  & 400 & 3000 \\ \hline
	\end{tabular}
\end{table}

\useunder{\uline}{\ul}{}
\begin{table}[]
	\centering
	\caption[]{Simulated anomaly examples}
	\label{table:anomaly_example}
	\begin{tabular}{|c|c|c|c|c|c|c|c|c|c|}
		\hline
		\multicolumn{2}{|c|}{\textbf{Example}} &
		\textbf{Fx} &
		\textbf{Dose/Fx} &
		\textbf{Age at Tx} &
		\textbf{Technique} &
		\textbf{Energy} &
		\textbf{Intent} &
		\textbf{ICD10 code} &
		\textbf{Morphology code} \\ \hline
		\multirow{2}{*}{A} &
		orig &
		5 &
		400 &
		76 &
		3D &
		mixed photon &
		- &
		C34.90 &
		80463 \\ \cline{2-10} 
		&
		mutate &
		{\ul \textbf{4}} &
		{\ul \textbf{500}} &
		76 &
		3D &
		mixed photon &
		- &
		C34.90 &
		80463 \\ \hline
		\multirow{2}{*}{B} &
		orig &
		5 &
		1000 &
		91 &
		SBRT &
		x06fff &
		curative &
		R91.1 &
		- \\ \cline{2-10} 
		&
		mutate &
		5 &
		1000 &
		{\ul \textbf{10}} &
		SBRT &
		x06fff &
		{\ul \textbf{palliative}} &
		R91.1 &
		- \\ \hline
		\multirow{2}{*}{C} &
		orig &
		4 &
		1200 &
		49 &
		SBRT &
		x06 &
		palliative &
		C34.30 &
		87203 \\ \cline{2-10} 
		&
		mutate &
		4 &
		1200 &
		49 &
		SBRT &
		{\ul \textbf{x10}} &
		palliative &
		{\ul \textbf{C15.9}} &
		87203 \\ \hline
		\multirow{2}{*}{D} &
		orig &
		10 &
		300 &
		74 &
		3D &
		x15 &
		palliative &
		C78.1 &
		- \\ \cline{2-10} 
		&
		mutate &
		10 &
		300 &
		74 &
		{\ul \textbf{IMRT}} &
		x15 &
		palliative &
		C78.1 &
		- \\ \hline
		\multicolumn{10}{|l|}{- stands for missing values} \\ \hline
	\end{tabular}
\end{table}

\clearpage

\begin{table}[H]
	\caption{Mock peer review examples (performance from physicians and model) }
	\label{table:cases}
	\centering
	\begin{tabular}{|ccccccccccc|}
		\hline
		\multicolumn{11}{|c|}{\textbf{Case 1}} \\ \hline
		\multicolumn{1}{|c|}{\multirow{3}{*}{}} &
		\multicolumn{1}{c|}{\textbf{Tx site}} &
		\multicolumn{1}{c|}{\textbf{Fx}} &
		\multicolumn{1}{c|}{\textbf{Dose/Fx}} &
		\multicolumn{1}{c|}{\textbf{Tech.}} &
		\multicolumn{1}{c|}{\textbf{Energy}} &
		\multicolumn{1}{c|}{\textbf{Tx   intent}} &
		\multicolumn{1}{c|}{\textbf{\begin{tabular}[c]{@{}c@{}}ICD10\\ code\end{tabular}}} &
		\multicolumn{1}{c|}{\textbf{\begin{tabular}[c]{@{}c@{}}Morph.\\ code\end{tabular}}} &
		\multicolumn{1}{c|}{\textbf{\begin{tabular}[c]{@{}c@{}}Age\\ at Tx\end{tabular}}} &
		\textbf{Truth} \\ \cline{2-11} 
		\multicolumn{1}{|c|}{} &
		\multicolumn{1}{c|}{\multirow{2}{*}{lul nodule}} &
		\multicolumn{1}{c|}{\multirow{2}{*}{10}} &
		\multicolumn{1}{c|}{\multirow{2}{*}{5000}} &
		\multicolumn{1}{c|}{\multirow{2}{*}{SBRT}} &
		\multicolumn{1}{c|}{\multirow{2}{*}{x6fff}} &
		\multicolumn{1}{c|}{\multirow{2}{*}{curative}} &
		\multicolumn{1}{c|}{\multirow{2}{*}{R91.1}} &
		\multicolumn{1}{c|}{\multirow{2}{*}{-}} &
		\multicolumn{1}{c|}{\multirow{2}{*}{73}} &
		\multirow{2}{*}{Anomaly} \\
		\multicolumn{1}{|c|}{} &
		\multicolumn{1}{c|}{} &
		\multicolumn{1}{c|}{} &
		\multicolumn{1}{c|}{} &
		\multicolumn{1}{c|}{} &
		\multicolumn{1}{c|}{} &
		\multicolumn{1}{c|}{} &
		\multicolumn{1}{c|}{} &
		\multicolumn{1}{c|}{} &
		\multicolumn{1}{c|}{} &
		\\ \hline
		\multicolumn{1}{|c|}{MD1} &
		\multicolumn{1}{c|}{Flag} &
		\multicolumn{9}{c|}{Dose/Fx error, unsafe} \\ \hline
		\multicolumn{1}{|c|}{MD2} &
		\multicolumn{1}{c|}{No Flag} &
		\multicolumn{9}{c|}{} \\ \hline
		\multicolumn{1}{|c|}{MD3} &
		\multicolumn{1}{c|}{Flag} &
		\multicolumn{9}{c|}{Prescription error} \\ \hline
		\multicolumn{1}{|c|}{Model} &
		\multicolumn{1}{c|}{Flag} &
		\multicolumn{9}{c|}{Type 1 anomaly. R = 1.982, $t_{Rx}$=0.002} \\ \hline
		\multicolumn{11}{|c|}{\textbf{Case 2}} \\ \hline
		\multicolumn{1}{|c|}{\multirow{3}{*}{}} &
		\multicolumn{1}{c|}{\textbf{Tx site}} &
		\multicolumn{1}{c|}{\textbf{Fx}} &
		\multicolumn{1}{c|}{\textbf{Dose/Fx}} &
		\multicolumn{1}{c|}{\textbf{Tech.}} &
		\multicolumn{1}{c|}{\textbf{Energy}} &
		\multicolumn{1}{c|}{\textbf{Tx   intent}} &
		\multicolumn{1}{c|}{\textbf{\begin{tabular}[c]{@{}c@{}}ICD10\\ code\end{tabular}}} &
		\multicolumn{1}{c|}{\textbf{\begin{tabular}[c]{@{}c@{}}Morph.\\ code\end{tabular}}} &
		\multicolumn{1}{c|}{\textbf{\begin{tabular}[c]{@{}c@{}}Age\\ at Tx\end{tabular}}} &
		\textbf{Truth} \\ \cline{2-11} 
		\multicolumn{1}{|c|}{} &
		\multicolumn{1}{c|}{\multirow{2}{*}{lul centr nodule}} &
		\multicolumn{1}{c|}{\multirow{2}{*}{5}} &
		\multicolumn{1}{c|}{\multirow{2}{*}{1000}} &
		\multicolumn{1}{c|}{\multirow{2}{*}{SBRT}} &
		\multicolumn{1}{c|}{\multirow{2}{*}{x6fff}} &
		\multicolumn{1}{c|}{\multirow{2}{*}{palliative}} &
		\multicolumn{1}{c|}{\multirow{2}{*}{R91.1}} &
		\multicolumn{1}{c|}{\multirow{2}{*}{-}} &
		\multicolumn{1}{c|}{\multirow{2}{*}{10}} &
		\multirow{2}{*}{Anomaly} \\
		\multicolumn{1}{|c|}{} &
		\multicolumn{1}{c|}{} &
		\multicolumn{1}{c|}{} &
		\multicolumn{1}{c|}{} &
		\multicolumn{1}{c|}{} &
		\multicolumn{1}{c|}{} &
		\multicolumn{1}{c|}{} &
		\multicolumn{1}{c|}{} &
		\multicolumn{1}{c|}{} &
		\multicolumn{1}{c|}{} &
		\\ \hline
		\multicolumn{1}{|c|}{MD1} &
		\multicolumn{1}{c|}{No Flag} &
		\multicolumn{9}{c|}{} \\ \hline
		\multicolumn{1}{|c|}{MD2} &
		\multicolumn{1}{c|}{Flag} &
		\multicolumn{9}{c|}{\begin{tabular}[c]{@{}c@{}}palliative not usually used to define lung SBRT;\\ age atypical for lung SBRT\end{tabular}} \\ \hline
		\multicolumn{1}{|c|}{MD3} &
		\multicolumn{1}{c|}{Flag} &
		\multicolumn{9}{c|}{\begin{tabular}[c]{@{}c@{}}seems odd for 10 yr-old to receive such dose, \\ but I don't know peds cases very well.\end{tabular}} \\ \hline
		\multicolumn{1}{|c|}{Model} &
		\multicolumn{1}{c|}{Flag} &
		\multicolumn{9}{c|}{Type 2 anomaly. F = 0.269, $t_F$ = 0.234} \\ \hline
		\multicolumn{11}{|c|}{\textbf{Case 3}} \\ \hline
		\multicolumn{1}{|c|}{\multirow{3}{*}{}} &
		\multicolumn{1}{c|}{\textbf{Tx site}} &
		\multicolumn{1}{c|}{\textbf{Fx}} &
		\multicolumn{1}{c|}{\textbf{Dose/Fx}} &
		\multicolumn{1}{c|}{\textbf{Tech.}} &
		\multicolumn{1}{c|}{\textbf{Energy}} &
		\multicolumn{1}{c|}{\textbf{Tx   intent}} &
		\multicolumn{1}{c|}{\textbf{\begin{tabular}[c]{@{}c@{}}ICD10\\ code\end{tabular}}} &
		\multicolumn{1}{c|}{\textbf{\begin{tabular}[c]{@{}c@{}}Morph.\\ code\end{tabular}}} &
		\multicolumn{1}{c|}{\textbf{\begin{tabular}[c]{@{}c@{}}Age\\ at Tx\end{tabular}}} &
		\textbf{Truth} \\ \cline{2-11} 
		\multicolumn{1}{|c|}{} &
		\multicolumn{1}{c|}{\multirow{2}{*}{left airways}} &
		\multicolumn{1}{c|}{\multirow{2}{*}{10}} &
		\multicolumn{1}{c|}{\multirow{2}{*}{300}} &
		\multicolumn{1}{c|}{\multirow{2}{*}{IMRT}} &
		\multicolumn{1}{c|}{\multirow{2}{*}{x15}} &
		\multicolumn{1}{c|}{\multirow{2}{*}{palliative}} &
		\multicolumn{1}{c|}{\multirow{2}{*}{C78.1}} &
		\multicolumn{1}{c|}{\multirow{2}{*}{-}} &
		\multicolumn{1}{c|}{\multirow{2}{*}{74}} &
		\multirow{2}{*}{Anomaly} \\
		\multicolumn{1}{|c|}{} &
		\multicolumn{1}{c|}{} &
		\multicolumn{1}{c|}{} &
		\multicolumn{1}{c|}{} &
		\multicolumn{1}{c|}{} &
		\multicolumn{1}{c|}{} &
		\multicolumn{1}{c|}{} &
		\multicolumn{1}{c|}{} &
		\multicolumn{1}{c|}{} &
		\multicolumn{1}{c|}{} &
		\\ \hline
		\multicolumn{1}{|c|}{MD1} &
		\multicolumn{1}{c|}{Flag} &
		\multicolumn{9}{c|}{Why IMRT used for palliative?} \\ \hline
		\multicolumn{1}{|c|}{MD2} &
		\multicolumn{1}{c|}{Flag} &
		\multicolumn{9}{c|}{IMRT for palliative dosing (300/10) is not typical} \\ \hline
		\multicolumn{1}{|c|}{MD3} &
		\multicolumn{1}{c|}{No Flag} &
		\multicolumn{9}{c|}{} \\ \hline
		\multicolumn{1}{|c|}{Model} &
		\multicolumn{1}{c|}{Flag} &
		\multicolumn{9}{c|}{Type 2 anomaly. F = 0.585, $t_F$ = 0.438} \\ \hline
		\multicolumn{11}{|l|}{- stands for missing values} \\ \hline
	\end{tabular}
\end{table}
\clearpage
In Supplementary Table S6, the first case is a simulated anomaly (SA), where the original record was 5 x 1000.
Both MD1 and MD3 identified it as an prescription anomaly, however, MD2 missed.
The model predicted it as a Type 1 anomaly because the evaluation metrics $R$ is greater than $t_{Rx}$. 

The second case is also a SA, because we changed the age from 91 to 10 and treatment intent from curative to palliative. 
The fact that the treatment intent for this Rx is never curative and never been given to a pediatric patient creates a mismatch between the feature set and the prescription, even though the Rx is one of the most popular Rx in historical SBRT database. MD2 seems to found both atypical properties of this patient while MD3 found one reason and MD1 missed it completely. 
The model predicted it as a Type 2 anomaly because the evaluation metrics $F$ is greater than $t_F$.  

The last case is also a SA, where we change the label from 3D to IMRT, that all of the features became mismatched with this prescription. MD1 and MD2 identified it as anomaly but MD3 missed it. The model predicted as a Type 2 anomaly because again $F$ is greater than $t_F$. 
\subsection{Supplementary figures}
\begin{figure}[H]
	\centering
	\includegraphics[width=0.7\textwidth]{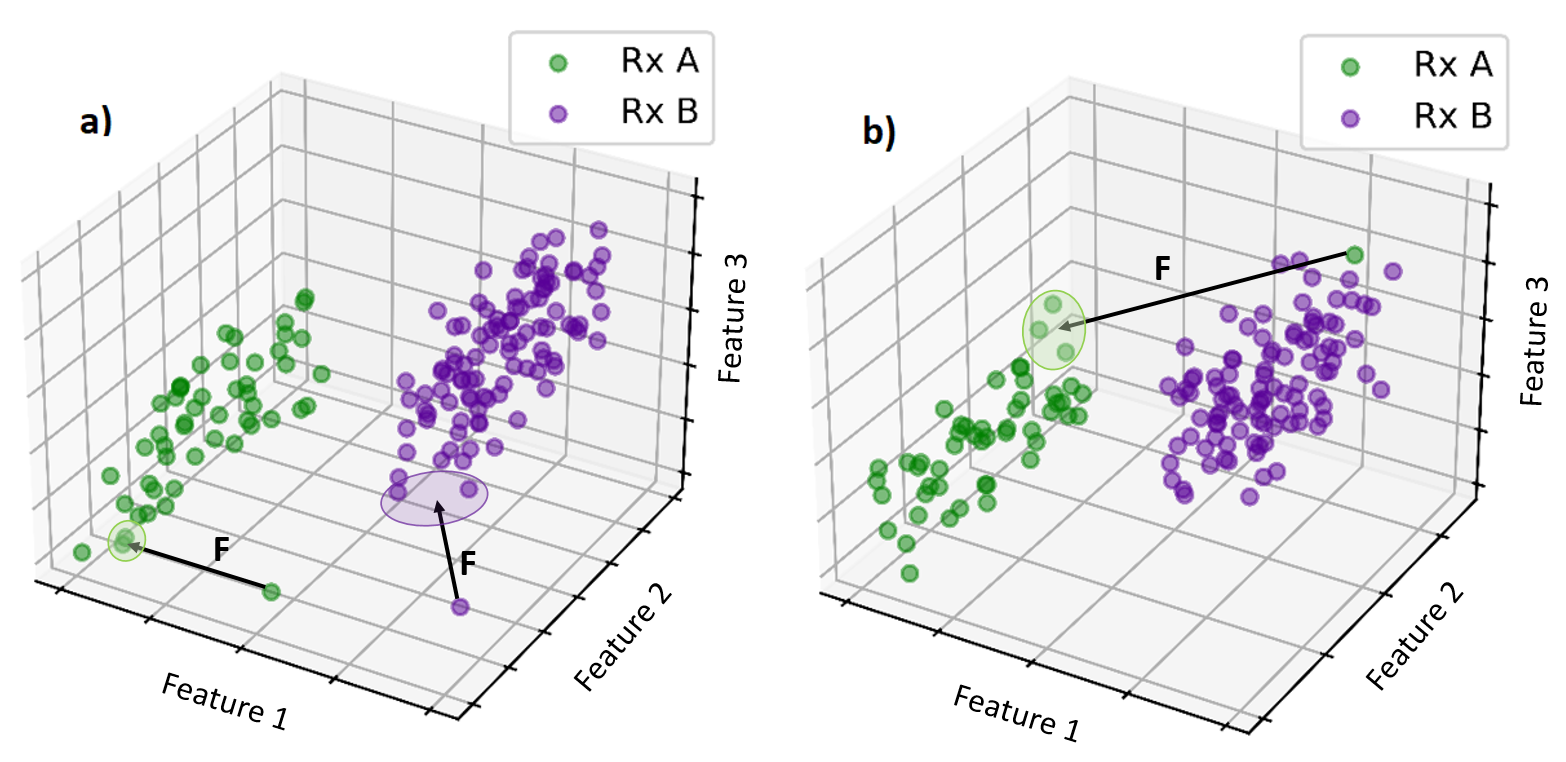}
	
	\caption{{\bf Illustration of different anomalous cases the model is designed to catch.} In {\bf a)}, we show two feature anomalies that are far from the average. In {\bf b)} a case with Rx A is mismatched within the feature sector of Rx B.} 
	\label{fig:3dplot}
\end{figure}

\begin{figure}[H]
	\centering
	\includegraphics[width=0.56\textwidth]{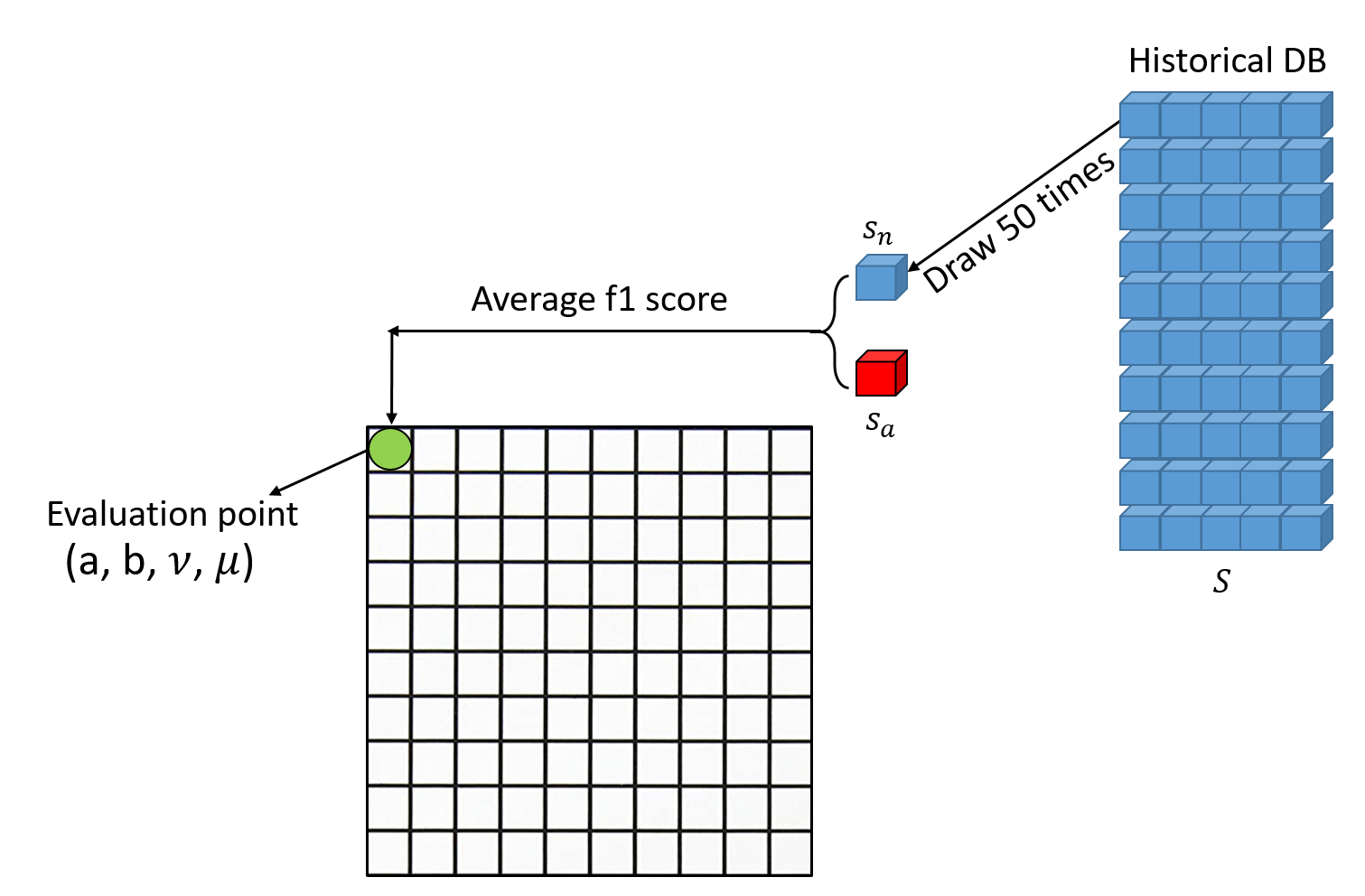}
	\caption{{\bf An illustration of the training process}. In each point in parameter grid space ($a, b, \nu, \mu$), we run the model 50 times, where for each run, normal samples (size of $s_{n}$) are randomly chosen from the historical database (size of $S$). In the meanwhile, anomaly samples (size of $s_{a}$) are kept the same. After the 50 runs, we will average $f1$ score and record the value for this evaluation point. Then, we move on to the next evaluation point. Eventually, 100 grid points are evaluated. Among those, we chose the parameter set that will give the highest $f1$ score. It is possible that this optimized parameter set is not unique (two grid points can provide the same highest $f1$ score).}
	\label{fig:algorithm}
\end{figure}

\begin{figure}[H]
	\centering
	\includegraphics[width=0.65\textwidth]{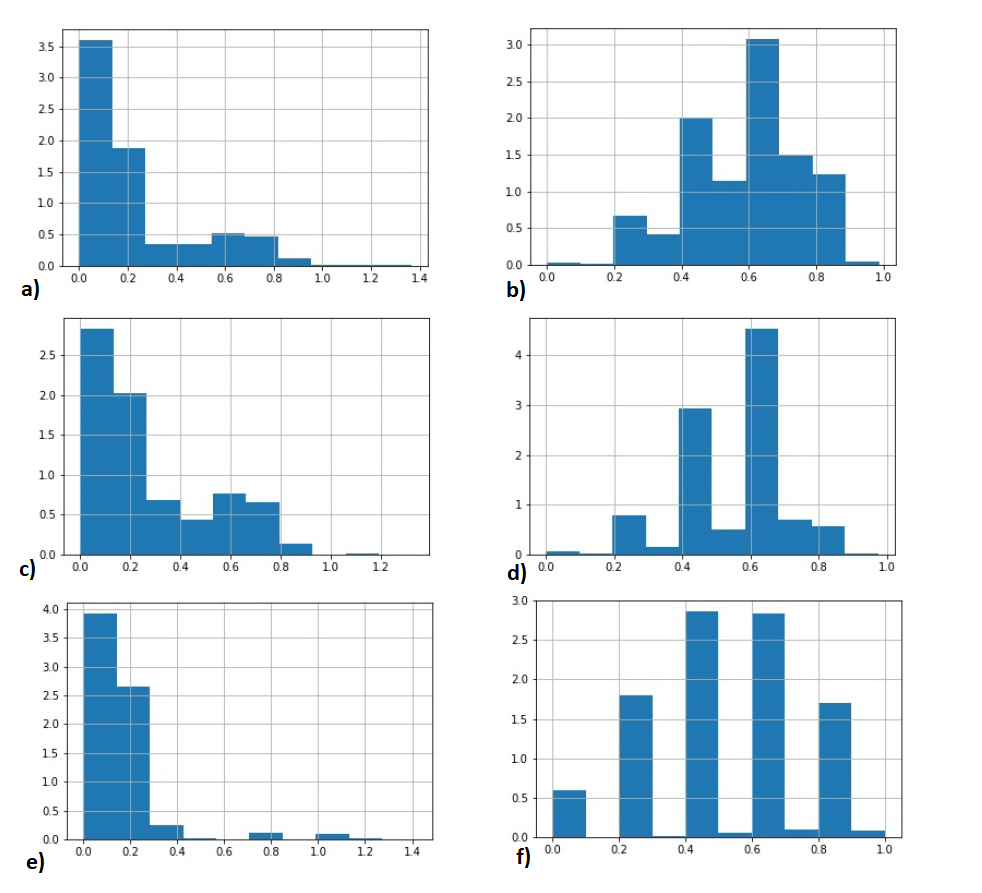}
	\caption{{\bf Normalized histograms of $Rx$ and feature distances in the historical patients' database.} Panel {\bf a)-b)} 3d {\bf c)-d)} IMRT {\bf e)-f)} SBRT. Left panel is the distribution of pairwise prescription distance $\rho_{Rx}(j,k)$ and right panel is distribution of Gower feature distance $g_{F}(j,k)$ with $j, k$ are historical patients' pair. Spikes relate to the three categorical features through the dice similarity. If one categorical feature is different the net dissimilarity is 0.2, if two categoricals are different it is 0.4 etc. In SBRT, the pairwise Gower distances are dominated by categorical features.}
	\label{fig:pairwise_dist}
\end{figure}

\end{document}